\documentclass{article}
\usepackage{ijcai26}

\usepackage{times}
\usepackage{soul}
\usepackage{url}
\usepackage[hidelinks]{hyperref}
\usepackage[utf8]{inputenc}
\usepackage[small]{caption}
\usepackage{graphicx}
\usepackage{amsmath}
\usepackage{amssymb}
\usepackage{amsthm}
\usepackage{booktabs}
\usepackage{algorithm}
\usepackage{algorithmic}
\usepackage[switch]{lineno}

\let\citep\cite
\newcommand{\citet}[1]{\citeauthor{#1}~(\citeyear{#1})}

\title{Evaluating RL Explainability Methods\\by How Much They Help Fix Bugs in Agents}

\author{
Ram Rachum$^{1,2}$
\and
Yotam Amitai$^{3}$\and
B\'{a}lint Gyevn\'{a}r$^{4}$\and
Reuth Mirsky$^{2}$\And
Cameron Allen$^{1}$\\
\affiliations
$^1$University of California, Berkeley\\
$^2$Tufts University\\
$^3$Independent Researcher\\
$^4$Carnegie Mellon University\\
\emails
ram.rachum@berkeley.edu,
yotamitai@gmail.com,
bgyevnar@cmu.edu,
reuth.mirsky@tufts.edu,
camallen@berkeley.edu
}

\usepackage[normalem]{ulem}
\usepackage{xcolor}
\usepackage{soul}% for strikethrough \st and underline \ul
\usepackage[normalem]{ulem}% for \sout and \uline that need to e.g. wrap \citet
\newcommand{\reviewersafecites}{%
  \let\reviewerorigcitet\citet
  \let\reviewerorigcitep\citep
  \let\reviewerorigcite\cite
  \renewcommand{\citet}[1]{\mbox{\reviewerorigcitet{##1}}}%
  \renewcommand{\citep}[1]{\mbox{\reviewerorigcitep{##1}}}%
  \renewcommand{\cite}[1]{\mbox{\reviewerorigcite{##1}}}%
}
\newcommand{\definereviewer}[2]{%
  \expandafter\newcommand\csname #1color\endcsname{#2}%
  \expandafter\newcommand\csname #1uline\endcsname{%
    \bgroup
    \markoverwith{\textcolor{#2}{\rule[-0.5ex]{2pt}{0.4pt}}}%
    \ULon}%
  \expandafter\DeclareRobustCommand\csname #1\endcsname[1]{%
    \textcolor{#2}{$\langle$##1$\rangle$}}%
  \expandafter\DeclareRobustCommand\csname #1ul\endcsname[2]{%
    \begingroup\reviewersafecites
      \csname #1uline\endcsname{##1}%
    \endgroup
    \csname #1\endcsname{##2}}%
  \expandafter\DeclareRobustCommand\csname #1del\endcsname[1]{%
    \begingroup\reviewersafecites
      \csname #1\endcsname{\sout{##1}}%
    \endgroup}%
  \expandafter\DeclareRobustCommand\csname #1chg\endcsname[2]{%
    \begingroup\reviewersafecites
      \csname #1\endcsname{\sout{##1}}%
    \endgroup
    \begingroup\reviewersafecites
      \csname #1uline\endcsname{##2}%
    \endgroup}%
}
\definereviewer{cam}{orange}

\begin{document}

\maketitle

\begin{abstract}
This preliminary paper outlines a planned evaluation benchmark for Explainable Reinforcement Learning (XRL) methods. Current evaluations rely on functionally-grounded metrics like faithfulness and compactness, and on human-grounded proxies like subjective ratings or prediction accuracy. We suggest evaluating XRL methods by how effectively their generated explanations help to diagnose and fix malfunctioning reinforcement learning (RL) agents. We propose \textbf{EvalXRL}, a benchmark in which a Large Language Model (LLM) coding agent uses different XRL methods to diagnose a held-out malfunction in an RL agent, and then repair it. Our proposed benchmark iterates across (environment $\times$ malfunction $\times$ XRL method) tuples and uses the reward signal of the RL agents to form a final score for each XRL method. The coding agent may use the method \textit{interactively}: invoke the XRL method, process its output, form new hypotheses on what is broken, and invoke the method again with parameters adjusted for testing these hypotheses. This closed-loop structure may be described as a simplified version of the scientific method. Some XRL methods provide self-evaluations that follow this pattern; we propose the first head-to-head comparison of multiple XRL methods in closed-loop usage.
\end{abstract}

%=======================================================
\section{Introduction}
\label{sec:intro}

XRL methods produce explanations that answer several distinct needs. Developers want to know why their agent is failing; end-users want to know whether an agent is trustworthy; regulators want accountability in systems that affect the public. All these stakeholders want an explanation, but each wants to get something different out of it, so each has different criteria for what they will consider a good explanation \citep{miller2019explanation}.

The landscape of XRL methods is as varied as the makeup of its target audience. Methods differ in what kind of questions they answer, in what information they use to build answers, and in the form their answers take. A saliency map \citep{greydanus2018visualizing} marks which pixels of an observation drove a single action; HIGHLIGHTS \citep{amir2018highlights} returns short video clips of the policy's most important moments; reward decomposition \citep{juozapaitis2019reward} reports per-component contributions to a scalar value; and VIPER \citep{bastani2018verifiable} distills the whole policy into a readable decision tree. Given this Cambrian explosion of outputs, the XRL community would benefit from evaluation methods that take in each of these heterogeneous artifacts and produce a homogeneous array of comparable scores. Hopefully, these scores would answer the questions ``how useful are the explanations produced by each method?'' and ``which methods are the most useful?'' keeping in mind that the different types of XRL users we described above might each require a different evaluation scheme.

Evaluating explanations is a difficult problem, because an explanation is considered good to the extent that it helps the user \textit{understand}; therefore, each XRL evaluation paradigm can be interpreted as an answer to the epistemological question ``what does it mean to \textit{understand}?'' \citep{lipton2009understanding,deRegt2017understanding} Table~\ref{tab:paradigms} surveys three existing paradigms for XRL evaluation, each framed as an implicit answer to this question.

\begin{table}[h]
\caption{Existing paradigms for XRL evaluation, framed as implicit answers to \textit{``what does it mean to understand?''}}
\label{tab:paradigms}
\centering
\small
\begin{tabular}{@{}p{0.42\linewidth}p{0.5\linewidth}@{}}
\toprule
\textbf{Paradigm} & \textbf{Implicit answer} \\
\midrule
Subjective satisfaction and trust ratings \citep{hoffman2019metrics} & ``If enough people feel that they understand, then they do.'' \\
\addlinespace
Fidelity and faithfulness measures \citep{xiong2024xrlbench} & ``To understand is to create a simpler model that produces similar results as a complex model.'' \\
\addlinespace
Simulatability tests \citep{hase2020evaluating,anderson2019explaining} & ``To understand is to be able to predict the outputs.'' \\
\bottomrule
\end{tabular}
\end{table}

We are proposing an evaluation paradigm whose respective answer is: \textbf{``to understand a mechanism is to be able to fix it when it breaks.''} This functional view of explanation is long-standing in cognitive science \citep{lombrozo2006}, with a recent ability-based formulation for AI \citep{chen2026machine}; as an evaluation paradigm it is partial like the others, but measurable. Therefore, our main focus is the first motivation we listed: the developer's need to know why their agents are failing. This is often called \textit{debugging}~\cite{gyevnar2025objective}, and we will use the terms debugging and \textit{diagnosis} synonymously. Diagnosing failures is an arduous but necessary phase in the endeavor of developing increasingly capable RL agents. In this paper, we propose a method to evaluate which XRL methods produce explanations that are most useful for diagnosis. The framing follows the spirit of \citet{doshivelez2017interpretable}'s \textit{application-grounded} evaluation, with the developer's downstream task as the metric. It echoes recent calls for downstream-validated, fair-fight interpretability evaluation \citep{marks2025downstream,nanda2025pragmatic} and for treating intervention outcomes as ground truth \citep{barez2026agenda}. To our knowledge, though, this has not been executed as a benchmark, and not for explainable RL.

This approach offers several benefits. First, we can wire the RL agent's reward signal directly to the output of our evaluation. In other words, we define the success of an XRL method as the performance of an RL agent that was repaired using information from that method. This unmediated connection may be a good strategy for removing proxy-driven bias from our scores. Most excitingly, diagnosis and repair are interactive processes, in which an explainability method may be used multiple times; we elaborate in design choice~2 below.

We propose \textbf{EvalXRL}, a benchmark for evaluating XRL methods by how well they help a developer repair online-RL agents. It operationalizes task completion as follows: (1) give the developer a deliberately broken RL agent and an XRL method; (2) ask them to repair it; (3) measure how well the repaired agent performs, as reported by its reward signal or a held-out scoring function. The score is always continuous, never a binary ``did you find the bug?'' judgment; we never ask the developer to articulate the cause, only to make the agent perform better. Three design choices follow:

\begin{enumerate}
    \item \textbf{Deliberately broken RL agents as evaluation substrate.} We construct agents with known, controlled malfunctions whose diagnosis requires reasoning about behavioral consequences. Holding the malfunction set fixed across all XRL methods lets us measure each method's diagnostic utility on like-for-like cells.
    \item \textbf{The developer may use the method multiple times, and adjust its parameters.} Diagnosis is a closed-loop process: the developer invokes the XRL method, processes its output, forms new hypotheses about what is broken, and re-invokes the method with parameters adjusted to test these specific hypotheses. This is in contrast to one-shot or pre-planned multi-shot use (open-loop), where parameters are fixed without reference to intermediate results. We make closed-loop use a first-class property of the evaluation: the repair score is computed after arbitrarily many XRL method invocations within the per-session budget, not after one. Some XRL methods provide self-evaluations in a closed-loop pattern \citep{amitai2024asqit,wu2022langxrl,dodge2021aar,cruz2021interactive}; we propose the first head-to-head comparison of multiple XRL methods in closed-loop usage.
    \item \textbf{We use an LLM coding agent as the developer.} Running our proposed procedure with LLM agents is orders of magnitude cheaper and faster than running it with human subjects, and avoids complex approvals for human research. In our default configuration the developer is a frontier LLM (e.g., Claude Opus~4.8) wrapped in a minimalist coding-agent scaffold (e.g., \textit{mini-swe-agent}~\citep{mini-swe-agent}), running in a sandboxed Docker environment with full programmatic control: it can read source, run experiments, modify parameters, retrain the agent, and submit a fix. We measure variance across $N$ runs per cell. We discuss the tradeoffs of more elaborate scaffolds in Section~\ref{sec:open}.
\end{enumerate}

A measurement of how effective an XRL method is for helping an LLM agent fix a bug may not carry over accurately enough to the use case of helping a human, and we sketch a calibration between the two in Section~\ref{sec:open}. However, we argue that besides serving as a proxy for humans, this measurement may be useful in and of itself. \citet{bhatt2020explainable} surveyed deployed XAI systems and found that the dominant user is the ML engineer who is debugging a model, not a layperson; and LLM coding agents are increasingly doing engineer-shaped work, with clear evidence that engineering of ML systems is being automated with LLM assistance~\citep{kulibaba2025Kompete,liu2025MLMasterAI}, often referred to as ``vibe coding''. This suggests that LLM coding agents may take a larger role in debugging work that was previously done by human engineers. If ``vibe debugging'' becomes prevalent, we advocate that it should be studied in controlled settings.

We present EvalXRL as a benchmark \textit{design} and a set of hypotheses, not yet as a validated leaderboard. The pilot run that would test H1--H3 is the immediate next step, and is the contribution of a follow-up paper. The rest of this paper describes the design, the hypotheses, and open questions we want the community's input on.

%=======================================================
\section{Related Work}
\label{sec:related}

The closest prior work to EvalXRL is \citet{gyevnar2025objective}, who divide XRL evaluation into two contexts: \textit{debugging} (developers verifying or investigating an agent before/after deployment) and \textit{teaming} (humans and agents collaborating during deployment). Their five debugging metrics are all prediction-based: next-action, goal, sub-goal, counterfactual policy, and time taken. Task completion appears in their layout only on the teaming side. The cell their layout does not contain is \textit{task completion as a debugging metric}: not predicting what the agent will do, but fixing what the broken agent fails to do and then measuring the fixed agent. EvalXRL is an attempt to fill that cell.

\paragraph{Cross-method XRL comparisons.} Recent work has compared multiple XRL methods on the same task with the same users via objective accuracy. \citet{septon2023integrating} crossed HIGHLIGHTS with reward decomposition on Highway-env and Pacman; \citet{amitai2024coviz} found counterfactual-outcome explanations and reward decomposition complementary on a highway agent; \citet{towers2025comparative} evaluated four mechanisms on Pacman; and \citet{frost2022counterfactual} compared counterfactual versus critical-state trajectories under distribution shift. EvalXRL extends this comparison style from prediction and identification tasks to repair tasks.

\paragraph{Repair as the XAI use case.} \citet{bhatt2020explainable} found that ML engineers debugging models are XAI's dominant real-world consumers. In supervised XAI, \citet{adebayo2020debugging} introduced \textit{debugging tests} that evaluate explanation methods against known model bugs; we adapt this framing to RL by scoring XRL methods on downstream repair success. The closest precursor to EvalXRL is \citet{sequeira2020interestingness}, who constructed Frogger agents with controlled deficits and ran a user study where participants identified each agent's capabilities and limitations from XRL-generated visual summaries. \citet{olson2021counterfactual} extended this to flaw-detection: non-experts who watched counterfactual states from a Space Invaders deep RL agent whose ship-position pixels had been masked raised flaw identification from 57\% to 90\%. Earlier, \citet{hayes2017improving} generated natural-language policy explanations to help operators pinpoint controller faults, and \citet{cruz2021interactive} had non-experts iteratively patch a Super Mario Bros agent with interactive explanations. \citet{amitai2024asqit}'s ASQ-IT lets users iteratively query an RL agent's behavior, with a user study showing that interactive querying improves users' ability to identify faulty behavior; this directly motivates EvalXRL's closed-loop design choice (Section~\ref{sec:intro}). None of these studies compare XRL methods head-to-head on a controlled malfunction set. \citet{paleja2021utility} ran the closest task-completion study, but with a hand-designed agent rather than a learned RL policy. \citet{tappler2025legible}'s LEGIBLE demonstrates that the diagnose-and-repair loop is already viable inside RL, lifting cumulative reward by up to 273\% on highway-fast without retraining, though as a single method it offers no head-to-head comparison.

\paragraph{Dialogic and hypothesis-driven XAI.} The argument that explanation is fundamentally dialogical rather than one-shot has been made repeatedly. \citet{miller2019explanation} reviews the cognitive-science evidence that explanation is an abductive-reasoning loop and a social process. \citet{madumal2018towards} derive a grounded dialog model from $398$ explanation dialogs in which follow-up questions and re-explanation are first-class moves. \citet{lakkaraju2022rethinking}'s interviews with $26$ domain experts (medical and policy researchers) report that all but one cited the impossibility of conversing with explanations as a primary frustration. \citet{miller2023evaluative} argues for a paradigm shift to hypothesis-driven decision support, what he calls \textit{evaluative AI}; \citet{le2024evidence} provide a Weight-of-Evidence implementation reporting improved trust calibration over a recommendation-driven baseline, with mixed accuracy effects across populations. \citet{wu2022langxrl}'s LangXRL operationalizes a related idea in RL: users edit history information and observe counterfactual changes in agent behavior, with larger gains on abnormality identification and downstream actionability against a static-text baseline. \citet{dodge2021aar} extend the structured-iteration argument with AAR/AI, an after-action-review protocol that frames RL agent assessment as Popperian falsification. EvalXRL sits in this lineage but inverts the human/machine roles: the LLM coder, not the human, drives the closed loop, which lets the comparison run at scale and be scored end-to-end on a downstream repair task.

\paragraph{LLM coding agents with tools.} A parallel thread asks whether giving an LLM coding agent a debugging or ML-engineering tool improves repair performance. \citet{yuan2025debuggym}'s Debug-Gym lifts SWE-bench Lite scores by exposing \texttt{pdb}, and \citet{levin2025chatdbg} let an LLM autonomously drive \texttt{pdb}/\texttt{gdb}/\texttt{lldb} to fix student Python bugs. \citet{haque2025execution} provide a counterpoint: naive injection of full execution traces helps in only 2 of 6 dataset-model configurations. \citet{chan2024mlebench}'s MLE-bench and \citet{wijk2024rebench}'s RE-Bench drop LLM agents into sandboxed ML engineering tasks scored by execution; EvalXRL inherits their harness pattern but re-targets it at XRL-method-assisted repair of malfunctioning RL agents. \citet{hariharan2025breakpoint}'s Breakpoint takes the closest methodological step: it auto-generates code-repair tasks by corrupting functions in real repositories, using fault-injection rather than naturally-occurring bugs. EvalXRL's malfunction-injection design follows the same logic, applied to RL.

\paragraph{LLM surrogates and auditing precedents.} \citet{debona2024evaluating} showed that LLMs can replicate human conclusions on XAI explanation evaluation tasks, while \citet{gao2025caution} caution that LLMs do not always replicate human behavior distributions. In XRL specifically, \citet{belouadah2025evaluating} used LLMs both to generate explanations and to judge explanation quality. \citet{mills2023almanacs}'s ALMANACS uses an LLM as an automated predictor and finds that no explanation method beats its no-explanation baseline, a sobering precedent for why we include a strong no-method floor. Closer to EvalXRL's design, \citet{marks2025auditing} introduced an auditing game in which research teams tried to uncover a deliberately-trained model's hidden objective, and \citet{bricken2025auditing} extended this to autonomous investigator agents. \citet{sheshadri2026auditbench}'s AuditBench scales the format and reports a \textit{tool-to-agent gap}: tools that score well in standalone evaluations often fail to help an agentic investigator. \citet{zhong2026pando}'s Pando reaches a similar empirical floor in alignment auditing: most white-box methods fail to outperform a budget-matched black-box prompting baseline. We borrow the auditing-game format and apply it to XRL by making an LLM coding agent the active surrogate user. The tool-to-agent gap is a direct warning about EvalXRL's methodology, which is part of why we include a strong no-method baseline.

\textbf{Gap.} The combination we plan to fill: a head-to-head, application-grounded benchmark for XRL methods, scored by repair-task completion of deliberately broken RL agents, with an LLM coding agent as the scalable surrogate user.

%=======================================================
\section{Proposed Framework}
\label{sec:framework}

\subsection{Terminology}
\label{subsec:terminology}

We use a small, deliberate vocabulary throughout. The \textbf{coder} is the agent acting as the surrogate user; in our default configuration it is an LLM coding agent, but the framework is agnostic and a human could in principle play the same role. The \textbf{RL agent} is the reinforcement learning agent the coder is trying to repair, often called \textit{broken} or \textit{malfunctioning} when it carries a deliberate fault. A \textbf{malfunction} is that fault, injected as a unified diff to the training code or environment. Each XRL \textbf{method} is packaged as a \textbf{dossier} containing a bundle of papers, code, documentation, and analysis tools that are added to the coder's context and command list. A \textbf{cell} is one (malfunction, method) pair, the unit of replication in the experimental design. The \textbf{repair score} is the post-repair RL agent's normalized task performance, in $[0, 1]$, computed by the harness, continuous (never binary), and averaged across $N$ replications per cell. The \textbf{harness} is the sandboxed Docker setup that runs each replication: two containers on an internal network with no internet access, one for the coder's workspace, one for the immutable scoring environment.

\subsection{Architecture}
\label{subsec:architecture}

The harness is a two-container Docker setup, both containers cut off from the internet. The first container holds the coder's workspace: the environment source code, the malfunctioning RL agent's parameters, the active XRL method, and an agent server that exposes the policy over the network. The coder has standard coding-agent tools (shell, Python, file editing) plus scoring and submission tools, and any additional tools the active method declares. The second container holds an immutable copy of the environment, drives episodes against the agent server, and computes the repair score. The coder cannot access or modify the second container, so scoring is tamper-proof. With no internet access, the coder also cannot look up solutions.

\subsection{Environments}

We plan to evaluate on a mix of environments where repairing the RL agent has a clear real-world analogue, all implemented in JAX \citep{jax2018github} with fully JIT-compiled training. Our current candidate set:

\begin{itemize}
    \item \textbf{Treasure Grid (TG)}: a $6\times6$ gridworld with heterogeneous reward tiles (large treasure $+10$, small treasure $+1$, pits $-5$).
    \item \textbf{Datacenter Cooling (DC)}: a simplified thermal model with server racks, cooling units, and an external weather model. Continuous high-dimensional observation, continuous action, multi-component reward (energy cost, thermal safety, equipment wear). The repair scenario maps to a technician confronting a malfunctioning cooling agent and trying to restore acceptable performance on energy-and-temperature metrics.
    \item \textbf{Traffic Light Control (TLC)}: a multi-intersection signalized traffic network. Discrete-action signal phase selection, per-intersection reward derived from standard transportation metrics (vehicle delay, throughput, queue length). The repair scenario maps to a transit engineer fixing an underperforming controller deployment.
\end{itemize}

\noindent We are also considering CartPole as a small control-task baseline, and applied alternatives such as energy grid balancing, building HVAC, inventory management, and packet routing. We welcome the community's suggestions.

\subsection{Malfunctions}

Each malfunction is injected via a unified diff applied to the training code or environment. The design draws on real-fault taxonomies for DRL \citep{nikanjam2022faults} and the RLMutation operator set \citep{tambon2023mutation}, extended to XRL-relevant failure modes such as reward hacking and non-stationary control. \citet{engstrom2020implementation} showed that subtle code-level details in PPO drive larger performance swings than the choice of algorithm itself, which validates that small "boring" bugs are the right target.

\paragraph{Candidate malfunctions.} Diagnosing these requires reasoning about behavioral consequences, not just code inspection. Several are instances of goal misgeneralization \citep{langosco2022goal}, where the agent retains capability out of distribution yet competently pursues an unintended proxy. This is where XRL is supposed to help.
\begin{itemize}
    \item \textbf{Reward clipping} on Treasure Grid: a widely-used technique \citep{mnih2015humanlevel} that destroys the distinction between large and small treasures.
    \item \textbf{Reward hacking} on Treasure Grid: a step penalty that incentivizes hovering near small treasures rather than traversing the grid (a pattern of the type studied by \citet{pan2022reward} and formalized by \citet{skalse2022defining}).
    \item \textbf{Mild myopia} on Treasure Grid: a discount factor of $0.3$ instead of $0.99$. The agent still cares about the future but undershoots distant rewards, settling for nearby small treasures.
    \item \textbf{Reward imbalance} on Treasure Grid: the large-treasure reward is cut from $10$ to $1.5$, making it locally rational to camp on small treasures rather than traverse to a large one. Distinct from reward clipping (which truncates during training) and reward hacking (which adds a perverse incentive); this is a pure value-tuning bug in the environment constants.
    \item \textbf{Distributional shift} on Datacenter Cooling: an agent trained on summer weather but deployed in winter. No code bug.
    \item \textbf{Sensor drift} on Datacenter Cooling: a temperature sensor reading $5$\textdegree C higher, injected into the observation function.
    \item \textbf{Reward hacking} on Datacenter Cooling: oscillating setpoints that minimize an instantaneous-energy term while violating multi-step thermal constraints (a reward-hacking pattern of the type catalogued by \citet{pan2022reward}).
    \item \textbf{Distributional shift} on Traffic Light Control: an agent trained at off-peak demand levels deployed at rush-hour demand. No code bug.
    \item \textbf{Reward hacking} on Traffic Light Control: oscillating signal phases that inflate per-second throughput while increasing total queue length.
\end{itemize}

\subsection{Methods and Controls}

The framework is agnostic to which XRL methods are packaged. Every cell of the benchmark is reported alongside two \textbf{controls} that bracket the comparison from below and above and together implement the ``fair fight'' framing of \citet{marks2025downstream}: specify the affordances each method receives, then require it to beat the best non-explanation baseline given the same affordances.

\begin{itemize}
    \item \textbf{No-method baseline}: the coder is told that no XRL method is available and must debug from source code, parameters, and the environment alone. This is the lower control against which every XRL method is measured; H3 in particular is an explicit comparison against it.
    \item \textbf{Cheat oracle}: the coder is given a natural-language description of the malfunction. A soft upper bound on what any XRL method could provide; the symmetric counterpart to the no-method baseline.
\end{itemize}

\noindent The substantive XRL methods we plan to package between these two controls are:

\begin{itemize}
    \item \textbf{Reward decomposition} \citep{juozapaitis2019reward}: classifies each reward by its source component, showing what the agent is optimizing for and what it is neglecting.
    \item \textbf{Counterfactual analysis} \citep{vanderwaa2018contrastive,wehner2024counterfactual}: along representative trajectories, contrasts the agent's actions or the resulting rewards against alternatives, surfacing what the agent is sensitive to.
    \item \textbf{Action explanations} \citep{gyevnar2026axis}: an LLM agent interrogates a counterfactual simulator and produces natural-language rationales (AXIS).
    \item \textbf{Saliency / attribution maps} \citep{greydanus2018visualizing,lundberg2017shap,beechey2023sverl}: per-feature importance scores, local for an individual decision or aggregated across the policy.
    \item \textbf{Decision-tree extraction} \citep{bastani2018verifiable}: distills the policy into a readable tree.
    \item \textbf{Programmatic policies} \citep{verma2018programmatically}: extracts a programmatic representation of the policy.
    \item \textbf{Behavior summarization} \citep{amir2018highlights}: surfaces the most informative trajectory snippets (HIGHLIGHTS).
\end{itemize}

\noindent We plan to start with the first three or four in the initial round and add others as resources allow. We welcome the community's suggestions for additional methods to package.

\subsection{Experimental Design}

The independent variable is which method the coder receives. The primary dependent variable is the \textbf{repair score}: the post-repair agent's task performance, normalized to $[0, 1]$ via $\mathrm{score} = \mathrm{clip}_{[0,1]}\!\left(\frac{P_{\text{post}} - P_{\text{broken}}}{P_{\text{clean}} - P_{\text{broken}}}\right)$, where $P_{\text{post}}$ is the post-repair agent's environment-specific performance, $P_{\text{broken}}$ is the unrepaired malfunctioning agent's performance, and $P_{\text{clean}}$ is the unmalfunctioned agent's performance (the natural reference, kept independent from the cheat oracle so the oracle remains a falsifiable upper bound rather than a definitional ceiling). The score is continuous, never binary. Throughout, a method's headline result is reported as its \emph{lift} over the no-method baseline on the same cell (its repair score minus the no-method coder's repair score), so that the coder's intrinsic debugging ability is differenced out and the benchmark credits a method only for the marginal value its explanations add. Because the lower clip would conflate harm with no-effect, we additionally report the un-clipped value $\widetilde{\mathrm{score}} = (P_{\text{post}} - P_{\text{broken}}) / (P_{\text{clean}} - P_{\text{broken}})$ as a secondary outcome, and use it as the primary outcome for H3 (where the direction of interest is below baseline). A secondary dependent variable is the number of tool calls before submission, as a coarse efficiency proxy. We considered a third dependent variable based on transcript analysis ("did the coder correctly identify the root cause") but expect such judgment to be subjective and brittle, so we keep it only as a qualitative annotation rather than a primary measure. Each cell is replicated $N \geq 6$ times; we will set the per-cell $N$ from a pilot variance estimate before the main round. Few-run RL evaluation is notoriously unstable \citep{henderson2018deep}, so we will report uncertainty intervals around per-cell means rather than point estimates only, following the recommendations of \citet{agarwal2021precipice}.

%=======================================================
\section{Hypotheses}
\label{sec:hypotheses}

We list three hypotheses about what the benchmark will measure. We deliberately do \emph{not} pre-specify which methods will help with which malfunctions; our hypotheses are about properties of the (method $\times$ malfunction) interaction structure.

\paragraph{H1 (interaction structure).} The matrix of mean repair scores across (method $\times$ malfunction) cells will exhibit non-trivial structure: methods will cluster into groups with similar success profiles across malfunctions, and malfunctions will cluster into groups for which similar methods help. We do not specify the clusters in advance: the test is whether the structure exists, not what shape it takes. Operationally, the (method $\times$ malfunction) interaction sum-of-squares from a two-way decomposition of per-cell repair scores will significantly exceed its null distribution under a permutation test that randomly permutes method labels within each malfunction (we will report effect-size estimates with stratified-bootstrap confidence intervals over replications, following \citet{agarwal2021precipice}). If true, this would mean XRL methods cannot be ranked on a single axis, and the field should adopt multi-condition evaluation suites rather than ranking on hand-picked benchmarks. The two ways H1 could fail are equally informative: a near-uniform matrix would indicate no method is useful across the board; a rank-1 matrix would indicate methods differ only in overall strength, not in what they expose.

\paragraph{H2 (oracle ceiling).} The cheat oracle will set a high empirical bound but will not always reach a perfect score, because implementing a fix can be harder than diagnosing it. We treat the oracle as an empirical, not a theoretical, ceiling: a high-quality XRL trace could surface fix-relevant information more directly than a natural-language bug description does, and an XRL method that exceeds the oracle on some cell would itself be an interesting finding rather than a contradiction. The size of the gap between the cheat oracle and an optimal repair will quantify how much of the difficulty is diagnosis versus engineering, which is itself a useful measurement.

\paragraph{H3 (methods can hurt).} For at least one (method, malfunction) cell, the XRL method will produce a mean repair score significantly below the no-method baseline, under a one-tailed paired comparison appropriate for the bounded outcome (e.g., Wilcoxon signed-rank or bootstrap difference-in-means). H3 is an existence-of-discovery claim: we therefore use Benjamini--Hochberg false-discovery-rate control at $q=0.05$ across all cells, rather than family-wise (Holm) correction, which would be unattainably strict for a grid of this size. We do not pre-specify which cells; the prediction is that the harm-effect exists somewhere. We expect this pattern to occur when a method confidently reports a true-but-not-causal symptom that misdirects the coder away from the actual failure mode. This mirrors the \textit{tool-to-agent gap} reported by \citet{sheshadri2026auditbench} in alignment auditing, and is consistent with \citet{lakkaraju2020fool}, who showed empirically that misleading explanations can manipulate user trust in tabular classification settings. It is the most surprising prediction we make: a sound interpretability tool can have negative downstream value when its output is well-formed but cause-irrelevant for the malfunction at hand. Falsifying H3 (no cell satisfies the threshold above) would be informative in the other direction, suggesting that XRL methods are at worst inert rather than misleading.

%=======================================================
\section{Open Design Questions}
\label{sec:open}

This section collects open questions in the EvalXRL design: some are limitations we acknowledge, some are extensions we plan to explore, and many are both. Each entry states the issue and our current thinking, and is offered as a place where community input would shape the eventual benchmark.

\paragraph{LLM-vs-human validity gap.} EvalXRL measures whether an explanation contains actionable information for an LLM coder, not for a human one. LLMs read structured text differently than humans read visualizations, and an XRL method that scores well with one need not score well with the other. The LLM coder is a fairly literal reader, so a method that fails the LLM test is unlikely to help a human practitioner, but the converse is not guaranteed. The natural calibration is a small (N${\sim}12$) semi-structured interview study comparing the coder's verdict on each cell against SWE-savvy human readers via inter-rater agreement (e.g.\ Cohen's $\kappa$), using the objective debugging measures of \citet{gyevnar2025objective} rather than repair task performance (the latter would confound explanation quality with the user's implementation skill). Thematic coding of interview transcripts would also surface qualitative failure modes the score alone would miss. We treat EvalXRL as a filter, not a substitute, until that calibration is run.

\paragraph{Closed-loop use is not unconditionally helpful.} Our design choice to allow closed-loop invocation reflects a longstanding argument for dialogical explanation, but the empirical case for closed-loop over one-shot use is mixed. \citet{amitai2024asqit}'s Study~2 found that participants using a non-interactive HIGHLIGHTS baseline produced more correct initial hypotheses than ASQ-IT participants did, with ASQ-IT's advantage emerging only after iterative verification. \citet{wu2022langxrl}'s discussion reports that users often did not know which parameters to start with when iteratively editing inputs. \citet{le2024evidence} report mixed accuracy effects from hypothesis-driven decision support across user populations. We adopt closed-loop use as a design choice on conceptual grounds and treat its actual benefit as an empirical question. A natural extension is to vary the per-session invocation budget down to a single invocation in the limit, which would isolate how much of each method's utility comes from closed-loop use versus one-shot consumption.

\paragraph{Method packaging.} The unit we actually evaluate is a dossier: a bundle of README, papers, source, and examples handed to the coder. Two competent dossiers for the same underlying method can differ in prose, examples, and visualization helpers, and could yield meaningfully different repair scores. A published EvalXRL number is therefore a property of (method $\times$ packaging), not of the method alone, so method authors could improve scores by improving packaging without improving the method. We mitigate by publishing every dossier and inviting authors to submit their preferred packaging. A direct probe of the confound would be a rich-versus-minimal dossier ablation pairing each method with two packaging variants of differing length and detail, which would quantify how much of the per-method score is attributable to packaging quality.

\paragraph{Scaffold choice.} Three candidate scaffolds bracket the design space. (i)~A \emph{minimalist scaffold} like \textit{mini-swe-agent}~\citep{mini-swe-agent} gives the LLM only a sandboxed shell, with no specialized tools beyond bash. This is the cleanest measurement instrument: most variance attaches to the LLM and to the XRL method (the variables we want to vary), not to scaffold cleverness; the choice also aligns with the method-minimalism principle of \citet{nanda2025pragmatic}, who advocate trying simpler tools before fancier ones. We currently treat this as the default. (ii)~An \emph{engineered Agent-Computer Interface} like \textit{SWE-agent}~\citep{yang2024sweagent} adds purpose-built tools (file viewer, structured editor, syntax linter) and lifts SWE-bench resolution by ${\sim}10$ percentage points over a vanilla shell baseline; the cost is framework lock-in to choices made for an earlier model era. (iii)~A \emph{self-evolving} scaffold like \textit{Live-SWE-agent}~\citep{xia2025liveswe} lets the agent synthesize its own ad-hoc tools per task; the empirical risk for EvalXRL is that the agent reinvents probing tools (saliency dumps, trajectory inspectors, value plotters) that overlap the function of the XRL method being evaluated, narrowing the measurable XRL contribution. We see no clean answer in advance: the measurement-cleanness vs.\ scaffold-capability tradeoff is exactly the kind of choice we expect community input to shape.

\paragraph{Synthetic malfunctions.} We suggest a hand-designed catalog of malfunctions (reward clipping, mild myopia, reward imbalance, distributional shift, sensor drift, reward hacking) rather than a sample of bugs that actually appear in deployed RL systems. The wild distribution would include data-pipeline errors, version drift between training and serving, RNG seed leaks, and emergent reward-hacking, none of which our synthetic catalog easily reproduces. Strong performance on EvalXRL is therefore evidence about utility on \emph{our} bug distribution, not the wild one. Sampling bugs from real RL incident reports or from the issue trackers of major RL libraries is a possible direction, though such bugs are often confounded with project-specific context that resists isolation.

\paragraph{Access model.} EvalXRL evaluates XRL methods under a white-box regime: the coder receives full source, weights, and the retraining loop. We treat this as the easier evaluation regime: a method that fails to help under full access is unlikely to help under the restricted black-box access of proprietary agents. A black-box variant would require a separate intervention API and a smaller applicable-method set (saliency, programmatic-policy extraction, and decision-tree distillation all require model internals).

\paragraph{Memorization and self-evaluation.} The sandbox blocks test-time leakage but not pretraining memorization: a frontier LLM's training corpus likely contains the bug archetypes, the XRL libraries, and the failure modes \citet{engstrom2020implementation} catalog. \citet{haklay2026pitfalls} confirm this concretely in the adjacent setting of mechanistic interpretability evaluation: Claude Opus~4.1, asked cold, can recite the exact attention-head indices and roles of the canonical Indirect Object Identification (IOI) circuit from memory, and the GPT-5 judge shows similar recall. Mitigations: (i)~a \textit{memorization probe} (no-method coder identifying the bug from a description alone, before environment access); (ii)~a held-out class of \textit{novel composite malfunctions} in the spirit of \citet{hariharan2025breakpoint}; and (iii)~reporting the no-method baseline so memorization-only ``improvements'' stay visible. Using one LLM family to evaluate methods consumed by the same family is also circular, so we plan a robustness check across a panel of additional model families.

\paragraph{One axis of explanation utility.} EvalXRL measures one thing: how much an explanation helps fix a malfunctioning agent. Real explanations serve other purposes too. They calibrate user trust, support accountability, build the user's mental model of the agent's reasoning, and communicate decisions to stakeholders. An explanation method could excel at any of these while contributing nothing to repair, and the converse also holds: a method that boosts our repair score might do so by short-circuiting understanding rather than building it. We do not claim repair-contribution is the most important purpose of an explanation, only that it is one important purpose, easy to measure, and currently absent from the field's standard evaluation toolbox.

\paragraph{Joint use of multiple methods.} The benchmark default gives the coder one method at a time. Real users may consult several methods simultaneously, and complementarity may matter, as \citet{amitai2024coviz} observed: counterfactual outcomes combined with reward decomposition outperform either alone. An extension equips the coder with multiple methods at once and measures each method's Shapley contribution to the repair score, isolating which method actually pulls weight in a given malfunction context.

\paragraph{Cross-environment composition.} Each environment in the current set is self-contained. A composed environment (e.g., a datacenter cooling agent that also handles dynamic pricing) would test how methods scale when a single agent is responsible for multiple coupled domains. The resulting interaction effects might surface different XRL strengths than the single-domain setup.

\paragraph{Generalization beyond reinforcement learning.} The same shape of evaluation transports to large language models. The coder, the dossier-vs-baseline structure, the repair-scored metric, and the two-container harness all carry over directly; what changes is the substrate (a deliberately malfunctioning open-source LLM at the 7--9B parameter scale) and the method space (mech-interp methods such as sparse autoencoder features, steering vectors, linear probes, activation and attribution patching, and the logit and tuned lenses, alongside matched-affordance behavioral baselines such as prompting, prefill, and chain-of-thought reading, which the current LLM-interpretability discourse \citep{nanda2025pragmatic,marks2025downstream} treats as first-class competitors). The malfunction catalog also shifts: published model-organism recipes (emergent misalignment, sleeper-agent backdoors, refusal-direction ablation, synthetic-document implanted beliefs, persona drift, eval-awareness training) produce controlled LoRA-scale faults that play the same role here as our per-environment injected diffs. The empirical regularity that simpler behavioral methods often match or beat mech-interp methods on adjacent tasks \citep{mills2023almanacs,sheshadri2026auditbench} gives this variant sharper baseline-controlled stakes than the RL setting allows; whether that pattern transfers to a repair-scored benchmark is the natural pilot for a separate paper.

%=======================================================
\section{Conclusion}
\label{sec:conclusion}

We propose EvalXRL, a benchmark that evaluates XRL methods by how well they help a coder diagnose and repair RL agents. If our hypotheses hold, EvalXRL will give the XRL community a common-test, low-cost way to measure the downstream utility of explanations, complementing the more expensive but more ecologically valid human studies the field cannot run at scale. As a workshop-stage proposal, we would especially value the community's feedback on (i)~whether our malfunction taxonomy covers the failure modes XRL practitioners actually care about, (ii)~whether the LLM-as-surrogate framing is a defensible first-pass methodology or whether the validity gap is too large to take useful conclusions from, and (iii)~which additional XRL methods the community would most want to see packaged in the first public release of the benchmark.

%=======================================================
\section*{Acknowledgments}

We thank our colleagues for helpful discussions and feedback: David Aha, Nitay Alon, J\'er\^ome Botoko Ekila, Eli Bronstein, David Manheim, and Yonatan Nakar.

%=======================================================
\bibliographystyle{named}
\bibliography{bibliography}

@inproceedings{gyevnar2025objective,
  title={Objective Metrics for Human-Subjects Evaluation in Explainable Reinforcement Learning},
  author={Gyevnar, Balint and Towers, Mark},
  booktitle={Multi-disciplinary Conference on Reinforcement Learning and Decision Making (RLDM)},
  year={2025},
  localpath={"$DXC/g/Gyevnar 2025, Objective Metrics for Human-Subjects Evaluation in Explainable Reinforcement Learning"}
}

@article{towers2025comparative,
  title={A Comparative User Evaluation of {XRL} Explanations using Goal Identification},
  author={Towers, Mark and Du, Yali and Freeman, Christopher and Norman, Timothy J.},
  journal={arXiv preprint arXiv:2510.16956},
  year={2025},
  localpath={"$DXC/t/Towers 2025, A Comparative User Evaluation of XRL Explanations using Goal Identification"}
}

@article{doshivelez2017interpretable,
  title={Towards A Rigorous Science of Interpretable Machine Learning},
  author={Doshi-Velez, Finale and Kim, Been},
  journal={arXiv preprint arXiv:1702.08608},
  year={2017},
  localpath={"$DXC/d/Doshi-Velez 2017, Towards A Rigorous Science of Interpretable Machine Learning"}
}

@misc{marks2025downstream,
  title={Downstream applications as validation of interpretability progress},
  author={Marks, Samuel},
  year={2025},
  howpublished={LessWrong / Alignment Forum},
  url={https://www.lesswrong.com/posts/wGRnzCFcowRCrpX4Y/downstream-applications-as-validation-of-interpretability}
}

@misc{nanda2025pragmatic,
  title={A Pragmatic Vision for Interpretability},
  author={Nanda, Neel and Engels, Joshua and Conmy, Arthur and Rajamanoharan, Senthooran and Chughtai, Bilal and McDougall, Callum and Kram{\'a}r, J{\'a}nos and Smith, Lewis},
  year={2025},
  howpublished={LessWrong / Alignment Forum},
  url={https://www.lesswrong.com/posts/StENzDcD3kpfGJssR/a-pragmatic-vision-for-interpretability},
  localpath={"$DXC/n/Nanda 2025, A Pragmatic Vision for Interpretability"}
}

@misc{barez2026agenda,
  title={Automated Interpretability-Driven Model Auditing and Control: A Research Agenda},
  author={Barez, Fazl and others},
  year={2026},
  howpublished={Oxford Martin AI Governance Initiative (AIGI)},
  url={https://aigi.ox.ac.uk/publications/automated-interpretability-driven-model-auditing-and-control-a-research-agenda},
  localpath={"$DXC/b/Barez 2026, Automated Interpretability-Driven Model Auditing and Control A Research Agenda"}
}

@article{lombrozo2006,
  title={The Structure and Function of Explanations},
  author={Lombrozo, Tania},
  journal={Trends in Cognitive Sciences},
  volume={10},
  number={10},
  pages={464--470},
  year={2006},
  publisher={Elsevier},
  doi={10.1016/j.tics.2006.08.004},
  localpath={"$DXC/l/Lombrozo 2006, The structure and function of explanations"}
}

@article{chen2026machine,
  title={Machine Understanding},
  author={Chen, Huili and Grimm, Stephen R. and Russakovsky, Olga and Lombrozo, Tania},
  journal={Trends in Cognitive Sciences},
  year={2026},
  issn={1364-6613},
  doi={10.1016/j.tics.2026.04.003},
  url={https://www.sciencedirect.com/science/article/pii/S136466132600077X},
  localpath={"$DXC/c/Chen 2026, Machine understanding"}
}

@misc{xiong2024xrlbench,
  title={{XRL-Bench}: A Benchmark for Evaluating and Comparing Explainable Reinforcement Learning Techniques},
  author={Xiong, Yu and Hu, Zhipeng and Huang, Ye and Wu, Runze and Guan, Kai and Fang, Xingchen and Jiang, Ji and Zhou, Tianze and Hu, Yujing and Liu, Haoyu and Lyu, Tangjie and Fan, Changjie},
  howpublished={arXiv preprint arXiv:2402.12685},
  year={2024},
  localpath={"$DXC/x/Xiong 2024, XRL-Bench - A Benchmark for Evaluating and Comparing Explainable Reinforcement Learning Techniques"}
}

@inproceedings{paleja2021utility,
  title={The Utility of Explainable {AI} in Ad Hoc Human-Machine Teaming},
  author={Paleja, Rohan and Ghuy, Muyleng and Arachchige, Nadun Ranawaka and Jensen, Reed and Gombolay, Matthew},
  booktitle={Advances in Neural Information Processing Systems},
  volume={34},
  year={2021},
  localpath={"$DXC/p/Paleja 2022, The Utility of Explainable AI in Ad Hoc Human-Machine Teaming"}
}

@inproceedings{cruz2021interactive,
  title={Interactive Explanations: Diagnosis and Repair of Reinforcement Learning Based Agent Behaviors},
  author={Arzate Cruz, Christian and Igarashi, Takeo},
  booktitle={IEEE Conference on Games (CoG)},
  pages={1--8},
  year={2021},
  publisher={IEEE},
  doi={10.1109/CoG52621.2021.9618999},
  localpath={"$DXC/a/Arzate Cruz 2021, Interactive Explanations Diagnosis and Repair of Reinforcement Learning Based Agent Behaviors"}
}

@inproceedings{anderson2019explaining,
  title={Explaining Reinforcement Learning to Mere Mortals: An Empirical Study},
  author={Anderson, Andrew and Dodge, Jonathan and Sadarangani, Amrita and Juozapaitis, Zoe and Newman, Evan and Irvine, Jed and Chattopadhyay, Souti and Fern, Alan and Burnett, Margaret},
  booktitle={International Joint Conference on Artificial Intelligence},
  pages={1328--1334},
  year={2019},
  localpath={"$DXC/a/Anderson 2019, Explaining RL to Mere Mortals"}
}

@inproceedings{debona2024evaluating,
  title={Evaluating Explanations Through {LLMs}: Beyond Traditional User Studies},
  author={De Bona, Francesco Bombassei and Dominici, Gabriele and Miller, Tim and Langheinrich, Marc and Gjoreski, Martin},
  booktitle={NeurIPS Workshop on Generative AI for Health},
  year={2024},
  localpath={"$DXC/d/De Bona 2024, Evaluating Explanations Through LLMs Beyond Traditional User Studies"}
}

@article{gao2025caution,
  title={Take caution in using {LLMs} as human surrogates},
  author={Gao, Yuan and Lee, Dokyun and Burtch, Gordon and Fazelpour, Sina},
  journal={Proceedings of the National Academy of Sciences},
  volume={122},
  number={24},
  pages={e2501660122},
  year={2025},
  doi={10.1073/pnas.2501660122},
  localpath={"$DXC/g/Gao 2025, Take Caution in Using LLMs as Human Surrogates - Scylla Ex Machina"}
}

@misc{kulibaba2025Kompete,
  title = {{{KompeteAI}}: {{Accelerated Autonomous Multi-Agent System}} for {{End-to-End Pipeline Generation}} for {{Machine Learning Problems}}},
  shorttitle = {{{KompeteAI}}},
  author = {Kulibaba, Stepan and Dzhalilov, Artem and Pakhomov, Roman and Svidchenko, Oleg and Gasnikov, Alexander and Shpilman, Aleksei},
  year = 2025,
  month = sep,
  number = {arXiv:2508.10177},
  eprint = {2508.10177},
  primaryclass = {cs},
  publisher = {arXiv},
  doi = {10.48550/arXiv.2508.10177},
  langid = {english},
  localpath={"$DXC/k/Kulibaba 2025, KompeteAI Accelerated Autonomous Multi-Agent System for End-to-End Pipeline Generation for Machine Learning Problems"}
}

@misc{liu2025MLMasterAI,
  title = {{{ML-Master}}: {{Towards AI-for-AI}} via {{Integration}} of {{Exploration}} and {{Reasoning}}},
  shorttitle = {{{ML-Master}}},
  author = {Liu, Zexi and Cai, Yuzhu and Zhu, Xinyu and Zheng, Yujie and Chen, Runkun and Wen, Ying and Wang, Yanfeng and E, Weinan and Chen, Siheng},
  year = 2025,
  month = jun,
  number = {arXiv:2506.16499},
  eprint = {2506.16499},
  primaryclass = {cs},
  publisher = {arXiv},
  doi = {10.48550/arXiv.2506.16499},
  langid = {english},
  localpath={"$DXC/l/Liu 2025, ML-Master Towards AI-for-AI via Integration of Exploration and Reasoning"}
}

@inproceedings{tappler2025legible,
  title={Rule-Guided Reinforcement Learning Policy Evaluation and Improvement},
  author={Tappler, Martin and Lopez-Miguel, Ignacio D. and Tschiatschek, Sebastian and Bartocci, Ezio},
  booktitle={International Joint Conference on Artificial Intelligence},
  year={2025},
  localpath={"$DXC/t/Tappler 2024, Rule-Guided Reinforcement Learning Policy Evaluation and Improvement"}
}

@inproceedings{adebayo2020debugging,
  title={Debugging Tests for Model Explanations},
  author={Adebayo, Julius and Muelly, Michael and Liccardi, Ilaria and Kim, Been},
  booktitle={Advances in Neural Information Processing Systems},
  volume={33},
  year={2020},
  localpath={"$DXC/a/Adebayo 2020, Debugging Tests for Model Explanations"}
}

@inproceedings{juozapaitis2019reward,
  title={Explainable Reinforcement Learning via Reward Decomposition},
  author={Juozapaitis, Zoe and Koul, Anurag and Fern, Alan and Erwig, Martin and Doshi-Velez, Finale},
  booktitle={IJCAI Workshop on Explainable Artificial Intelligence (XAI)},
  year={2019},
  localpath={"$DXC/j/Juozapaitis 2019, Explainable Reinforcement Learning via Reward Decomposition"}
}

@inproceedings{beechey2023sverl,
  title={Explaining Reinforcement Learning with Shapley Values},
  author={Beechey, Daniel and Smith, Thomas M. S. and {\c{S}}im{\c{s}}ek, {\"O}zg{\"u}r},
  booktitle={International Conference on Machine Learning},
  pages={2003--2014},
  year={2023},
  organization={PMLR},
  localpath={"$DXC/b/Beechey 2023, Explaining Reinforcement Learning with Shapley Values"}
}

@inproceedings{bastani2018verifiable,
  title={Verifiable Reinforcement Learning via Policy Extraction},
  author={Bastani, Osbert and Pu, Yewen and Solar-Lezama, Armando},
  booktitle={Advances in Neural Information Processing Systems},
  volume={31},
  year={2018},
  localpath={"$DXC/b/Bastani 2018, Verifiable Reinforcement Learning via Policy Extraction"}
}

@inproceedings{yang2024sweagent,
  title={{SWE-agent}: Agent-Computer Interfaces Enable Automated Software Engineering},
  author={Yang, John and Jimenez, Carlos E. and Wettig, Alexander and Lieret, Kilian and Yao, Shunyu and Narasimhan, Karthik and Press, Ofir},
  booktitle={Advances in Neural Information Processing Systems (NeurIPS)},
  year={2024},
  eprint={2405.15793},
  archivePrefix={arXiv},
  localpath={"$DXC/y/Yang 2024, SWE-agent Agent-Computer Interfaces Enable Automated Software Engineering"}
}

@article{yuan2025debuggym,
  title={{debug-gym}: A Text-Based Environment for Interactive Debugging},
  author={Yuan, Xingdi and Moss, Morgane M and El Feghali, Charbel and Singh, Chinmay and Moldavskaya, Darya and MacPhee, Drew and Caccia, Lucas and Pereira, Matheus and Kim, Minseon and Sordoni, Alessandro and C{\^o}t{\'e}, Marc-Alexandre},
  journal={arXiv preprint arXiv:2503.21557},
  year={2025},
  localpath={"$DXC/y/Yuan 2025, debug-gym A Text-Based Environment for Interactive Debugging"}
}

@article{levin2025chatdbg,
  title={{ChatDBG}: Augmenting Debugging with Large Language Models},
  author={Levin, Kyla H. and van Kempen, Nicolas and Berger, Emery D. and Freund, Stephen N.},
  journal={Proceedings of the ACM on Software Engineering},
  volume={2},
  number={FSE},
  articleno={FSE085},
  year={2025},
  doi={10.1145/3729355},
  localpath={"$DXC/l/Levin 2024, ChatDBG Augmenting Debugging with Large Language Models"}
}

@inproceedings{haque2025execution,
  title={Towards Effectively Leveraging Execution Traces for Program Repair with Code {LLMs}},
  author={Haque, Mirazul and Babkin, Petr and Farmahinifarahani, Farima and Veloso, Manuela},
  booktitle={Proceedings of the 4th International Workshop on Knowledge-Augmented Methods for Natural Language Processing (KnowledgeNLP)},
  pages={160--179},
  year={2025},
  publisher={Association for Computational Linguistics},
  localpath={"$DXC/h/Haque 2025, Towards Effectively Leveraging Execution Traces for Program Repair with Code LLMs"}
}

@inproceedings{amitai2024coviz,
  title={Explaining Reinforcement Learning Agents Through Counterfactual Action Outcomes},
  author={Amitai, Yotam and Septon, Yael and Amir, Ofra},
  booktitle={AAAI Conference on Artificial Intelligence},
  volume={38},
  pages={10003--10011},
  year={2024},
  doi={10.1609/aaai.v38i9.28863},
  localpath={"$DXC/a/Amitai 2024, Explaining Reinforcement Learning Agents through Counterfactual Action Outcomes"}
}

@article{amitai2024asqit,
  title={{ASQ-IT}: Interactive Explanations for Reinforcement-Learning Agents},
  author={Amitai, Yotam and Amir, Ofra and Avni, Guy},
  journal={Artificial Intelligence},
  volume={335},
  pages={104182},
  year={2024},
  publisher={Elsevier},
  localpath={"$DXC/a/Amitai 2024, ASQ-IT - Interactive Explanations for Reinforcement-Learning Agents"}
}

@inproceedings{septon2023integrating,
  title={Integrating Policy Summaries with Reward Decomposition for Explaining Reinforcement Learning Agents},
  author={Septon, Yael and Huber, Tobias and Andr{\'e}, Elisabeth and Amir, Ofra},
  booktitle={International Conference on Practical Applications of Agents and Multi-Agent Systems (PAAMS)},
  pages={320--332},
  year={2023},
  publisher={Springer},
  doi={10.1007/978-3-031-37616-0_27},
  localpath={"$DXC/s/Septon 2023, Integrating Policy Summaries with Reward Decomposition"}
}

@article{mnih2015humanlevel,
  title={Human-level control through deep reinforcement learning},
  author={Mnih, Volodymyr and Kavukcuoglu, Koray and Silver, David and Rusu, Andrei A. and Veness, Joel and Bellemare, Marc G. and Graves, Alex and Riedmiller, Martin and Fidjeland, Andreas K. and Ostrovski, Georg and others},
  journal={Nature},
  volume={518},
  number={7540},
  pages={529--533},
  year={2015},
  publisher={Nature Publishing Group},
  localpath={"$DXC/m/Mnih 2015, Human-Level Control through Deep RL"}
}

@misc{jax2018github,
  author = {Bradbury, James and Frostig, Roy and Hawkins, Peter and Johnson, Matthew James and Katariya, Yash and Leary, Chris and Maclaurin, Dougal and Necula, George and Paszke, Adam and Vander{P}las, Jake and Wanderman-{M}ilne, Skye and Zhang, Qiao},
  title = {{JAX}: composable transformations of {P}ython+{N}um{P}y programs},
  url = {http://github.com/jax-ml/jax},
  year = {2018}
}

@article{miller2019explanation,
  title={Explanation in Artificial Intelligence: Insights from the Social Sciences},
  author={Miller, Tim},
  journal={Artificial Intelligence},
  volume={267},
  pages={1--38},
  year={2019},
  publisher={Elsevier},
  localpath={"$DXC/m/Miller 2019, Explanation in Artificial Intelligence - Insights from the Social Sciences"}
}

@article{belouadah2025evaluating,
  title={Evaluating the Effectiveness of {LLMs} for Explainable Deep Reinforcement Learning},
  author={Belouadah, Ayoub and Ruiz-Rodr{\'\i}guez, Marcelo Luis and Kubler, Sylvain and Le Traon, Yves},
  journal={Machine Learning with Applications},
  volume={22},
  pages={100795},
  year={2025},
  publisher={Elsevier},
  localpath={"$DXC/b/Belouadah 2025, Evaluating the Effectiveness of LLMs for Explainable Deep Reinforcement Learning"}
}

@inproceedings{wehner2024counterfactual,
  title={Explaining Learned Reward Functions with Counterfactual Trajectories},
  author={Wehner, Jan and Oliehoek, Frans A. and Cavalcante Siebert, Luciano},
  booktitle={ECAI Workshop on Implementing AI Ethics through a Behavioural Lens (AIEB)},
  year={2024},
  localpath={"$DXC/w/Wehner 2024, Explaining Learned Reward Functions with Counterfactual Trajectories"}
}

@inproceedings{gyevnar2026axis,
  title={Integrating Counterfactual Simulations with Language Models for Explaining Multi-Agent Behaviour},
  author={Gyevnar, B{\'a}lint and Lucas, Christopher G. and Albrecht, Stefano V. and Cohen, Shay B.},
  booktitle={International Conference on Autonomous Agents and Multiagent Systems},
  year={2026},
  localpath={"$DXC/g/Gyevnar 2025, Integrating Counterfactual Simulations with Language Models for Explaining Multi-Agent Behaviour"}
}

@article{nikanjam2022faults,
  title={Faults in Deep Reinforcement Learning Programs: A Taxonomy and a Detection Approach},
  author={Nikanjam, Amin and Morovati, Mohammad Mehdi and Khomh, Foutse and Ben Braiek, Houssem},
  journal={Automated Software Engineering},
  volume={29},
  number={8},
  year={2022},
  publisher={Springer},
  localpath={"$DXC/n/Nikanjam 2021, Faults in deep reinforcement learning programs a taxonomy and a detection tool"}
}

@inproceedings{pan2022reward,
  title={The Effects of Reward Misspecification: Mapping and Mitigating Misaligned Models},
  author={Pan, Alexander and Bhatia, Kush and Steinhardt, Jacob},
  booktitle={International Conference on Learning Representations},
  year={2022},
  localpath={"$DXC/p/Pan 2022, Effects of Reward Misspecification"}
}

@inproceedings{skalse2022defining,
  title={Defining and Characterizing Reward Hacking},
  author={Skalse, Joar and Howe, Nikolaus H. R. and Krasheninnikov, Dmitrii and Krueger, David},
  booktitle={Advances in Neural Information Processing Systems},
  year={2022},
  eprint={2209.13085},
  archivePrefix={arXiv},
  localpath={"$DXC/s/Skalse 2022, Defining and Characterizing Reward Hacking"}
}

@article{hoffman2019metrics,
  title={Metrics for Explainable {AI}: Challenges and Prospects},
  author={Hoffman, Robert R. and Mueller, Shane T. and Klein, Gary and Litman, Jordan},
  journal={arXiv preprint arXiv:1812.04608},
  year={2018},
  localpath={"$DXC/h/Hoffman 2019, Metrics for Explainable AI"}
}

@inproceedings{bhatt2020explainable,
  title={Explainable Machine Learning in Deployment},
  author={Bhatt, Umang and Xiang, Alice and Sharma, Shubham and Weller, Adrian and Taly, Ankur and Jia, Yunhan and Ghosh, Joydeep and Puri, Ruchir and Moura, Jos{\'e} M. F. and Eckersley, Peter},
  booktitle={ACM Conference on Fairness, Accountability, and Transparency},
  pages={648--657},
  year={2020},
  doi={10.1145/3351095.3375624},
  localpath={"$DXC/b/Bhatt 2019, Explainable machine learning in deployment"}
}

@inproceedings{lundberg2017shap,
  title={A Unified Approach to Interpreting Model Predictions},
  author={Lundberg, Scott M. and Lee, Su-In},
  booktitle={Advances in Neural Information Processing Systems},
  year={2017},
  localpath={"$DXC/l/Lundberg 2017, A Unified Approach to Interpreting Model Predictions"}
}

@inproceedings{greydanus2018visualizing,
  title={Visualizing and Understanding {Atari} Agents},
  author={Greydanus, Sam and Koul, Anurag and Dodge, Jonathan and Fern, Alan},
  booktitle={International Conference on Machine Learning},
  year={2018},
  localpath={"$DXC/g/Greydanus 2018, Visualizing and Understanding Atari Agents"}
}

@inproceedings{vanderwaa2018contrastive,
  title={Contrastive Explanations for Reinforcement Learning in terms of Expected Consequences},
  author={van der Waa, Jasper and van Diggelen, Jurriaan and van den Bosch, Karel and Neerincx, Mark},
  booktitle={IJCAI Workshop on Explainable Artificial Intelligence (XAI)},
  year={2018},
  localpath={"$DXC/v/van der Waa 2018, Contrastive Explanations for Reinforcement Learning in terms of Expected Consequences"}
}

@inproceedings{verma2018programmatically,
  title={Programmatically Interpretable Reinforcement Learning},
  author={Verma, Abhinav and Murali, Vijayaraghavan and Singh, Rishabh and Kohli, Pushmeet and Chaudhuri, Swarat},
  booktitle={International Conference on Machine Learning},
  year={2018},
  localpath={"$DXC/v/Verma 2018, Programmatically Interpretable Reinforcement Learning"}
}

@inproceedings{amir2018highlights,
  title={{HIGHLIGHTS}: Summarizing Agent Behavior to People},
  author={Amir, Dan and Amir, Ofra},
  booktitle={International Conference on Autonomous Agents and Multiagent Systems},
  year={2018},
  localpath={"$DXC/a/Amir 2018, HIGHLIGHTS - Summarizing Agent Behavior to People"}
}

@inproceedings{hayes2017improving,
  title={Improving Robot Controller Transparency Through Autonomous Policy Explanation},
  author={Hayes, Bradley and Shah, Julie A.},
  booktitle={ACM/IEEE International Conference on Human-Robot Interaction (HRI)},
  pages={303--312},
  year={2017},
  doi={10.1145/2909824.3020233},
  localpath={"$DXC/h/Hayes 2017, Improving Robot Controller Transparency Through Autonomous Policy Explanation"}
}

@article{sequeira2020interestingness,
  title={Interestingness elements for explainable reinforcement learning: Understanding agents' capabilities and limitations},
  author={Sequeira, Pedro and Gervasio, Melinda},
  journal={Artificial Intelligence},
  volume={288},
  pages={103367},
  year={2020},
  localpath={"$DXC/s/Sequeira 2019, Interestingness Elements for Explainable Reinforcement Learning through Introspection"}
}

@article{olson2021counterfactual,
  title={Counterfactual State Explanations for Reinforcement Learning Agents via Generative Deep Learning},
  author={Olson, Matthew L. and Khanna, Roli and Neal, Lawrence and Li, Fuxin and Wong, Weng-Keen},
  journal={Artificial Intelligence},
  volume={295},
  pages={103455},
  year={2021},
  localpath={"$DXC/o/Olson 2021, Counterfactual State Explanations for Reinforcement Learning Agents via Generative Deep Learning"}
}

@inproceedings{hase2020evaluating,
  title={Evaluating Explainable {AI}: Which Algorithmic Explanations Help Users Predict Model Behavior?},
  author={Hase, Peter and Bansal, Mohit},
  booktitle={Annual Meeting of the Association for Computational Linguistics (ACL)},
  pages={5540--5552},
  year={2020},
  localpath={"$DXC/h/Hase 2020, Evaluating Explainable AI Which Algorithmic Explanations Help Users Predict Model Behavior"}
}

@inproceedings{frost2022counterfactual,
  title={Explaining Reinforcement Learning Policies through Counterfactual Trajectories},
  author={Frost, Julius and Watkins, Olivia and Weiner, Eric and Abbeel, Pieter and Darrell, Trevor and Plummer, Bryan and Saenko, Kate},
  booktitle={ICML Workshop on Human in the Loop Learning (HILL)},
  year={2021},
  localpath={"$DXC/f/Frost 2022, Explaining RL Policies through Counterfactual Trajectories"}
}

@article{marks2025auditing,
  title={Auditing language models for hidden objectives},
  author={Marks, Samuel and Treutlein, Johannes and Bricken, Trenton and Lindsey, Jack and Marcus, Jonathan and Mishra-Sharma, Siddharth and Ziegler, Daniel and others},
  journal={arXiv preprint arXiv:2503.10965},
  year={2025},
  localpath={"$DXC/m/Marks 2025, Auditing language models for hidden objectives"}
}

@inproceedings{henderson2018deep,
  title={Deep Reinforcement Learning that Matters},
  author={Henderson, Peter and Islam, Riashat and Bachman, Philip and Pineau, Joelle and Precup, Doina and Meger, David},
  booktitle={AAAI Conference on Artificial Intelligence},
  year={2018},
  localpath={"$DXC/h/Henderson 2018, Deep Reinforcement Learning That Matters"}
}

@inproceedings{engstrom2020implementation,
  title={Implementation Matters in Deep Policy Gradients: A Case Study on {PPO} and {TRPO}},
  author={Engstrom, Logan and Ilyas, Andrew and Santurkar, Shibani and Tsipras, Dimitris and Janoos, Firdaus and Rudolph, Larry and Madry, Aleksander},
  booktitle={International Conference on Learning Representations (ICLR)},
  year={2020},
  localpath={"$DXC/e/Engstrom 2020, Implementation Matters in Deep Policy Gradients - A Case Study on PPO and TRPO"}
}

@inproceedings{tambon2023mutation,
  title={Mutation Testing of Deep Reinforcement Learning Based on Real Faults},
  author={Tambon, Florian and Majdinasab, Vahid and Nikanjam, Amin and Khomh, Foutse and Antoniol, Giuliano},
  booktitle={IEEE International Conference on Software Testing, Verification and Validation (ICST)},
  year={2023},
  localpath={"$DXC/t/Tambon 2023, Mutation Testing of Deep Reinforcement Learning Based on Real Faults"}
}

@article{chan2024mlebench,
  title={{MLE}-bench: Evaluating Machine Learning Agents on Machine Learning Engineering},
  author={Chan, Jun Shern and Chowdhury, Neil and Jaffe, Oliver and Aung, James and Sherburn, Dane and Mays, Evan and Starace, Giulio and Liu, Kevin and Maksin, Leon and Patwardhan, Tejal and Weng, Lilian and Madry, Aleksander},
  journal={arXiv preprint arXiv:2410.07095},
  year={2024},
  note={ICLR 2025},
  localpath={"$DXC/c/Chan 2024, MLE-bench Evaluating Machine Learning Agents on Machine Learning Engineering"}
}

@article{wijk2024rebench,
  title={{RE-Bench}: Evaluating frontier {AI} {R\&D} capabilities of language model agents against human experts},
  author={Wijk, Hjalmar and Lin, Tao and Becker, Joel and Jawhar, Sami and Parikh, Neev and Broadley, Thomas and Chan, Lawrence and Chen, Michael and Clymer, Josh and Dhyani, Jai and others},
  journal={arXiv preprint arXiv:2411.15114},
  year={2024},
  localpath={"$DXC/w/Wijk 2024, RE-Bench Evaluating frontier AI R&D capabilities of language model agents against human experts"}
}

@inproceedings{langosco2022goal,
  title={Goal Misgeneralization in Deep Reinforcement Learning},
  author={Langosco, Lauro and Koch, Jack and Sharkey, Lee D. and Pfau, Jacob and Krueger, David},
  booktitle={International Conference on Machine Learning (ICML)},
  pages={12004--12019},
  volume={162},
  series={Proceedings of Machine Learning Research},
  publisher={PMLR},
  year={2022},
  localpath={"$DXC/l/Langosco 2021, Goal Misgeneralization in Deep Reinforcement Learning"}
}

@article{sheshadri2026auditbench,
  title={{AuditBench}: Evaluating Alignment Auditing Techniques on Models with Hidden Behaviors},
  author={Sheshadri, Abhay and Ewart, Aidan and Fronsdal, Kai and Gupta, Isha and Bowman, Samuel R. and Price, Sara and Marks, Samuel and Wang, Rowan},
  journal={arXiv preprint arXiv:2602.22755},
  year={2026},
  localpath={"$DXC/s/Sheshadri 2026, AuditBench Evaluating Alignment Auditing Techniques on Models with Hidden Behaviors"}
}

@misc{bricken2025auditing,
  title={Building and Evaluating Alignment Auditing Agents},
  author={Bricken, Trenton and Wang, Rowan and Bowman, Sam and Ong, Euan and Treutlein, Johannes and Wu, Jeff and Hubinger, Evan and Marks, Samuel},
  year={2025},
  howpublished={Anthropic Alignment Science Blog},
  url={https://alignment.anthropic.com/2025/automated-auditing/},
  localpath={"$DXC/b/Bricken 2025, Building and Evaluating Alignment Auditing Agents"}
}

@article{hariharan2025breakpoint,
  title={{Breakpoint}: Scalable evaluation of system-level reasoning in {LLM} code agents},
  author={Hariharan, Kaivalya and Girit, Uzay and Wang, Atticus and Andreas, Jacob},
  journal={arXiv preprint arXiv:2506.00172},
  year={2025},
  localpath={"$DXC/h/Hariharan 2025, Breakpoint Scalable evaluation of system-level reasoning in LLM code agents"}
}

@article{mills2023almanacs,
  title={{ALMANACS}: A Simulatability Benchmark for Language Model Explainability},
  author={Mills, Edmund and Su, Shiye and Russell, Stuart and Emmons, Scott},
  journal={arXiv preprint arXiv:2312.12747},
  year={2023},
  localpath={"$DXC/m/Mills 2023, ALMANACS A Simulatability Benchmark for Language Model Explainability"}
}

@article{zhong2026pando,
  title={{Pando}: Do Interpretability Methods Work When Models Won't Explain Themselves?},
  author={Zhong, Ziqian and Muhamed, Aashiq and Diab, Mona T. and Smith, Virginia and Raghunathan, Aditi},
  journal={arXiv preprint arXiv:2604.11061},
  year={2026},
  localpath={"$DXC/z/Zhong 2026, Pando - Do Interpretability Methods Work When Models Won't Explain Themselves"}
}

@article{haklay2026pitfalls,
  title={Pitfalls in Evaluating Interpretability Agents},
  author={Haklay, Tal and Prakash, Nikhil and Pandey, Sana and Torralba, Antonio and Mueller, Aaron and Andreas, Jacob and Rott Shaham, Tamar and Belinkov, Yonatan},
  journal={arXiv preprint arXiv:2603.20101},
  year={2026},
  localpath={"$DXC/h/Haklay 2026, Pitfalls in Evaluating Interpretability Agents"}
}

@book{deRegt2017understanding,
  title={Understanding Scientific Understanding},
  author={de Regt, Henk W.},
  year={2017},
  publisher={Oxford University Press},
  doi={10.1093/oso/9780190652913.001.0001}
}

@incollection{lipton2009understanding,
  title={Understanding Without Explanation},
  author={Lipton, Peter},
  booktitle={Scientific Understanding: Philosophical Perspectives},
  editor={de Regt, Henk W. and Leonelli, Sabina and Eigner, Kai},
  pages={43--63},
  year={2009},
  publisher={University of Pittsburgh Press}
}

@article{lakkaraju2022rethinking,
  title={Rethinking Explainability as a Dialogue: A Practitioner's Perspective},
  author={Lakkaraju, Himabindu and Slack, Dylan and Chen, Yuxin and Tan, Chenhao and Singh, Sameer},
  journal={arXiv preprint arXiv:2202.01875},
  year={2022},
  localpath={"$DXC/l/Lakkaraju 2022, Rethinking Explainability as a Dialogue - A Practitioner's Perspective"}
}

@article{madumal2018towards,
  title={Towards a Grounded Dialog Model for Explainable Artificial Intelligence},
  author={Madumal, Prashan and Miller, Tim and Vetere, Frank and Sonenberg, Liz},
  journal={arXiv preprint arXiv:1806.08055},
  year={2018},
  localpath={"$DXC/m/Madumal 2018, Towards a Grounded Dialog Model for Explainable Artificial Intelligence"}
}

@inproceedings{lakkaraju2020fool,
  title={``How do {I} fool you?'': Manipulating User Trust via Misleading Black Box Explanations},
  author={Lakkaraju, Himabindu and Bastani, Osbert},
  booktitle={Proceedings of the AAAI/ACM Conference on AI, Ethics, and Society},
  pages={79--85},
  year={2020},
  localpath={"$DXC/l/Lakkaraju 2020, How do I Fool You - Manipulating User Trust via Misleading Black Box Explanations"}
}

@inproceedings{agarwal2021precipice,
  title={Deep Reinforcement Learning at the Edge of the Statistical Precipice},
  author={Agarwal, Rishabh and Schwarzer, Max and Castro, Pablo Samuel and Courville, Aaron and Bellemare, Marc G},
  booktitle={Advances in Neural Information Processing Systems (NeurIPS)},
  year={2021},
  localpath={"$DXC/a/Agarwal 2021, Deep Reinforcement Learning at the Edge of the Statistical Precipice"}
}

@article{wu2022langxrl,
  title={Decisions that Explain Themselves: A User-Centric Deep Reinforcement Learning Explanation System},
  author={Wu, Xiaoran and Yan, Zihan and Zhang, Chongjie and Wu, Tongshuang},
  journal={arXiv preprint arXiv:2212.00888},
  year={2022},
  localpath={"$DXC/w/Wu 2022, Decisions that Explain Themselves"}
}

@article{dodge2021aar,
  title={After-Action Review for {AI} ({AAR/AI})},
  author={Dodge, Jonathan and Khanna, Roli and Irvine, Jed and Lam, Kin-Ho and Mai, Theresa and Lin, Zhengxian and Kiddle, Nicholas and Newman, Evan and Anderson, Andrew and Raja, Sai and Matthews, Caleb and Perdriau, Christopher and Burnett, Margaret and Fern, Alan},
  journal={ACM Transactions on Interactive Intelligent Systems},
  year={2021},
  publisher={ACM},
  localpath={"$DXC/k/Khanna 2022, Finding AIs Faults with AAR-AI An Empirical Study"}
}

@inproceedings{miller2023evaluative,
  title={Explainable {AI} is Dead, Long Live Explainable {AI}! Hypothesis-Driven Decision Support Using Evaluative {AI}},
  author={Miller, Tim},
  booktitle={Proceedings of the 2023 ACM Conference on Fairness, Accountability, and Transparency (FAccT)},
  year={2023},
  localpath={"$DXC/m/Miller 2023, Explainable AI is Dead Long Live Explainable AI"}
}

@article{le2024evidence,
  title={From Evidence to Decision: Exploring Evaluative {AI}},
  author={Le, Thao and Miller, Tim and Sonenberg, Liz and Singh, Ronal and Soyer, H. Peter},
  journal={arXiv preprint arXiv:2402.01292},
  year={2024},
  localpath={"$DXC/l/Le 2024, From Evidence to Decision_ Exploring Evaluative AI"}
}

@article{xia2025liveswe,
  title={Live-{SWE}-agent: Can Software Engineering Agents Self-Evolve on the Fly?},
  author={Xia, Chunqiu Steven and others},
  journal={arXiv preprint},
  year={2025},
  localpath={"$DXC/x/Xia 2025, Live-SWE-agent Can Software Engineering Agents Self-Evolve on the Fly"}
}

@misc{mini-swe-agent,
  title={mini-swe-agent},
  author={{SWE-agent Team}},
  howpublished={\url{https://github.com/SWE-agent/mini-swe-agent}},
  year={2025},
  note={Software repository}
}

\end{document}